# Ego-Lane Analysis System (ELAS): Dataset and Algorithms

Rodrigo F. Berriel*, Edilson de Aguiar, Alberto F. de Souza, Thiago Oliveira-Santos

*Universidade Federal do Espírito Santo, Brazil*

**Abstract**

Decreasing costs of vision sensors and advances in embedded hardware boosted lane related research detection, estimation, tracking, etc. in the past two decades. The interest in this topic has increased even more with the demand for advanced driver assistance systems (ADAS) and self-driving cars. Although extensively studied independently, there is still need for studies that propose a combined solution for the multiple problems related to the ego-lane, such as lane departure warning (LDW), lane change detection, lane marking type (LMT) classification, road markings detection and classification, and detection of adjacent lanes (i.e., immediate left and right lanes) presence. In this paper, we propose a real-time Ego-Lane Analysis System (ELAS) capable of estimating ego-lane position, classifying LMTs and road markings, performing LDW and detecting lane change events. The proposed vision-based system works on a temporal sequence of images. Lane marking features are extracted in perspective and Inverse Perspective Mapping (IPM) images that are combined to increase robustness. The final estimated lane is modeled as a spline using a combination of methods (Hough lines with Kalman filter and spline with particle filter). Based on the estimated lane, all other events are detected. To validate ELAS and cover the lack of lane datasets in the literature, a new dataset with more than 20 different scenes (in more than 15,000 frames) and considering a variety of scenarios (urban road, highways, traffic, shadows, etc.) was created. The dataset was manually annotated and made publicly available to enable evaluation of several events that are of interest for the research community (i.e., lane estimation, change, and centering; road markings; intersections; LMTs; crosswalks and adjacent lanes). Moreover, the system was also validated quantitatively and qualitatively on other public datasets. ELAS achieved high detection rates in all real-world events and proved to be ready for real-time applications.

## 1. Introduction

Decreasing costs of vision sensors and advances in embedded hardware boosted traffic lane related research (detection, estimation, tracking, etc.) in the past two decades. The interest increased even more with the demand for advanced driver assistance systems (ADAS) and self-driving cars. The need for these solutions is also supported by the fact that humans are the main cause of car accidents [1]. Lane detection is an essential task in this context and has been extensively studied [2, 3, 4]. Nevertheless, drivers rely not only on the position of the lanes for driving safely, but also use visual cues (e.g., pavement markings) to understand what is and what is not allowed (direction, lane change, etc.) in a given lane.

Ego-lane and host lane are names given to the lane where the vehicle is positioned. Ego-lane analysis comprises multiple tasks related to the host lane, such as lane estimation (LE) [5], lane departure warning (LDW) [6], lane change detection [7], lane marking type (LMT) classification [8], road markings detection and classification [9], and detection of adjacent lanes, also known as multiple lanes detection [10, 11], etc. In general, different sensors have been used to address these issues: monocular [12] and stereo [13] cameras, LiDAR [14], and Sensor-fusion [15]. Each sensor has its own drawbacks (e.g., sudden illumination changes for cameras; sparsity and range limit for laser).

Many camera-based solutions [2, 10, 16] focus only on lane detection and estimation. Given the associated cost in generating datasets for these tasks, the ones used in many of the solutions in the literature are kept private. Alongside the lack of public datasets, there is a lack of public implementations which hinders fair comparisons. Although extensively studied independently, there is still need for studies that propose a combined real-time solution for the multiple problems related to the ego-lane including validation, public or open source implementation and public datasets, enabling fair comparisons.

In this context, this paper presents a real-time vision-based Ego-Lane Analysis System (ELAS) capable of estimating ego-lane position, classifying LMTs and road markings, performing LDW, detecting adjacent lanes (i.e., immediate left and right lanes), and detecting lane change events. The proposed monocular vision-based system works on a temporal sequence of images. The final estimated lane is modeled as a spline using a combination of methods (Hough lines with Kalman filter and spline with particle filter). Based on the estimated lane, all other events are detected using image processing techniques. A novel dataset was created and annotated. Alongside this dataset, the implementation of ELAS is publicly available for future fair comparisons. The dataset contains more than 20 different scenes (over 15,000 frames) and a variety of scenarios (urban road, highways, traffic, shadows, rain, etc.). Moreover, ELAS was evaluated on this dataset, achieving high detection rates in all real scenarios, proving to be ready for real-time real-world applications.

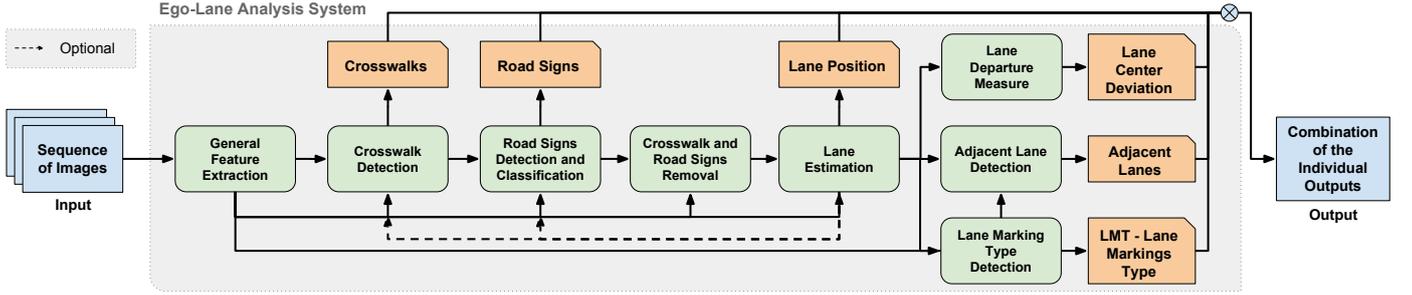

Figure 1: Overview of the Ego-Lane Analysis System (ELAS)

## 2. Ego-Lane Analysis System (ELAS)

ELAS processes a temporal sequence of images, analyzing each of them individually. These images are expected to come from a monocular forward-looking camera mounted on a vehicle. The general work-flow is described in Figure 1. Firstly, general road markings features are extracted and stored in feature maps. Based on these features, pavement markings (i.e., crosswalks, arrow markings, etc.) are detected and removed from the maps. Subsequently, lanes are estimated by using a combination of Hough lines with Kalman filter and spline with particle filter. These two processes are performed separately: the Hough lines with Kalman filter are used to estimate the base of the lane (i.e., base point, lane width, and lane direction); and, the spline-based particle filter is used to estimate the curvature of the lane. Essentially, the lane base estimated by the Hough lines and Kalman filter is used as a starting point to the particle filter, reducing the freedom of the spline near the car and guiding the spline curvature direction in the far depth of view. This combination was used to take advantage of the lane stability and linearity near the car (Hough and Kalman), while reducing the number of parameters of the particle filter responsible for the curvature. Finally, based on the estimated lane position, the remaining tasks are performed: LMT classification, adjacent lanes detection, and deviation from the lane center. All these tasks are performed in real-time (i.e., more than 30 frames per second) and are explained in detail in the following subsections. As a result, for each image the system outputs information describing the lane position, lane markings type, crosswalks, road signs, presence of adjacent lanes, and deviation of the car related to the lane center. It is worth mentioning that some of the algorithms hereby presented were designed considering the driving rules from Brazil (e.g., lane marking type, road signs, etc.). Many of these rules extend to other countries in the world, and therefore the assumptions and limitations would still be valid. In case of different rules, some adaptations of the presented method would be required.

### 2.1. General Feature Extraction

A general set of feature maps is extracted from each frame and the maps are used by each module according to its task. An overview of this process is shown in Figure 2.

#### 2.1.1. Preprocessing

Before generating the feature maps, the input image is converted to grayscale, since only pixel intensities are analyzed for generating the feature maps. Secondly, a region of interest (RoI) is set in order to remove irrelevant parts of the image. Finally, an Inverse Perspective Mapping (IPM) [17] is applied in order to reduce perspective distortion, resulting in a top-view image (also called bird's-eye view). To apply the IPM, ELAS assumes a constant ground plane along the input frames, using a static homography matrix. In this bird's-eye view image, lane markings tend to have constant width. However, due to road inclination and casual bumps, lane markings width constancy can be temporarily lost because fixed ground plane assumption may fail. As explained later, this issue is addressed by the lane model used.

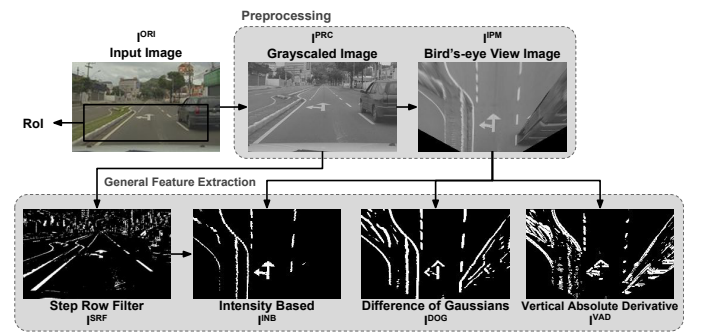

Figure 2: Overview of General Feature Extraction. Input image is preprocessed and features are extracted in the form of four binary maps, i.e., feature maps.

### 2.1.2. Feature Maps Generation

Four feature maps are generated using threshold-based techniques. Each feature map is a binary image. White pixels in these maps are called evidences. An overview of this process is described in Figure 2.

*Step Row Filter Map ($I^{SRF}$).* Lane markings are usually areas brighter than its surroundings (asphalt). In this feature map, lane marking evidences are detected on the original grayscaled image using the step row filter defined by Equation 1, as presented in [18].

$$y_i = 2x_i - (x_{i-\tau} + x_{i+\tau}) - |x_{i-\tau} - x_{i+\tau}| \qquad (1)$$

Here, $\tau$ is assumed large enough to surpass the size of double lane markings. Additionally, this filter was applied in the original image, linearly adjusting $\tau$ according to the vertical axis.

*Horizontal Difference of Gaussians Map ($I^{DOG}$).* Based on the same assumption (lane markings are brighter than asphalt), in this map, evidences are found using a horizontal Difference of Gaussians [19] (DoG), equivalent to a horizontal Mexican hat with central opening with the average size of a lane width. DoG is applied in the IPM grayscale image in order to take advantage of lane width invariability, and is followed by a threshold.

*Vertical Absolute Derivate Map ($I^{VAD}$).* Some objects of interest (e.g., vehicles, stop lines, etc.) contain horizontal edges in the bird's-eye view image. To extract these features vertical changes are calculated using the absolute value of the y-image derivative followed by a threshold.

*Intensity Based Map ($I^{INB}$).* This map aims to enhance the lane markings detection in $I^{SRF}$ by analyzing the intensity of the detected evidences. Therefore, it receives the $I^{SRF}$ and the grayscale input image as input. The inverse of $I^{SRF}$ is assumed to be composed of asphalt and others. Hence, the mean ($\mu_A$) and standard deviation ($\sigma_A$) in the grayscaled input image considering pixels that are not evidences in $I^{SRF}$ are used to filter wrongly detected lane markings in the original $I^{SRF}$. This process generates an intermediary feature map ($I^{INB'}$) comprising detected evidences in $I^{SRF}$ that have corresponding intensities in the grayscale input image with values higher than $\mu_A + 2\sigma_A$. The same procedure is applied in the $I^{INB'}$ in order to find out an optimal threshold for the lane markings. Therefore, the mean ($\mu_{LM}$) and standard deviation ($\sigma_{LM}$) in the grayscale input image considering pixels that are evidences in $I^{INB'}$ are used to calculate a better threshold for the lane markings in the grayscale input image, i.e.,

$I^{INB}$ comprises pixels in the grayscale input image with intensity values higher than $\mu_{LM} - \sigma_{LM}$. This process is summarized in equation Equation 2

$$\begin{aligned}
I^A &= \neg(I^{PRC} \wedge I^{SRF}) \\
\mu_A &= \overline{I^A}, \sigma_A = \sqrt{\overline{I^A - \mu_A}} \\
I^{INB'}_{x,y} &= \begin{cases} 1 & \text{if } (I^{PRC}_{x,y} > \mu_A + 2\sigma_A) \wedge (I^{SRF} > 0) \\ 0 & \text{otherwise} \end{cases} \\
\mu_{LM} &= \overline{I^{INB'}}, \sigma_{LM} = \sqrt{\overline{I^{INB'} - \mu_{LM}}} \\
I^{INB}_{x,y} &= \begin{cases} 1 & \text{if } I^{PRC}_{x,y} > \mu_{LM} - \sigma_{LM} \\ 0 & \text{otherwise} \end{cases}
\end{aligned} \qquad (2)$$

## 2.2. Crosswalk Detection

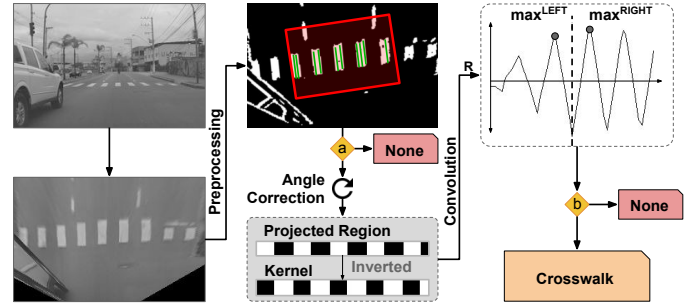

Figure 3: Overview of Crosswalk Detection. $I^{DOG}$ is analyzed in order to detect crosswalk candidates in the frame. A region of interest (red) is used to search for Houghs (green). If a set of Hough lines is detected, the region is rotated using the angle of the Houghs. The final crosswalk detection is based on the analyses of the convolution between the projected crosswalk region and its inverted version.

Crosswalks are series of large vertical strips (Figure 5) to mark a place where pedestrians can safely cross a street. Therefore, the goal of this module is detect crosswalks ahead of the vehicle. It uses the DoG-based feature map ($I^{DOG}$) as input to emphasize the vertical strips, and it takes advantage from the previous Lane Position (LP), if available. Firstly, a morphological closing is applied to remove small holes in $I^{DOG}$. Subsequently, a morphological erode is applied in order to degrade road markings that are expected to have smaller width than crosswalks. These preprocessing operations are based on the assumption that crosswalks are usually uniform in the IPM image. After that, a predefined region of the frame that is located in front of the vehicle is chosen as a crosswalk searching region. If a previous frame LP is available, lane direction is considered to properly define the crosswalk searching region according to the lane orientation. In order to detect crosswalks, features are extracted from this region using a Hough Transform. Further crosswalk detection processing is only carried on if Hough lines are detected, otherwise the region is assumed to have no crosswalk (Figure 3 (a)). Extracted Hough lines are grouped into a 2-dimensional histogram (see Figure 4) considering their angle and intersection with the bottom of the IPM image (with bin sizes equal to 1 and 3, respectively). This 2D histogram is denoted as $H_{x,y}^{2D}$, where $0 \leq y < M$ and $0 \leq x < N$ (in this case, $M = 360$ and $N = 640$). It is assumed that the angle of the crosswalk in relation to the image can be calculated by the dominant angle of the extracted Hough lines. As crosswalks could be rotated in relation to the car or lane itself, region is rotated in the direction of the dominant angle $\alpha$ of the extracted Hough lines. The dominant angle $\alpha$ is calculated (Figure 4) as the maximum number of hits considering the sum of a region along the axis of the angles:

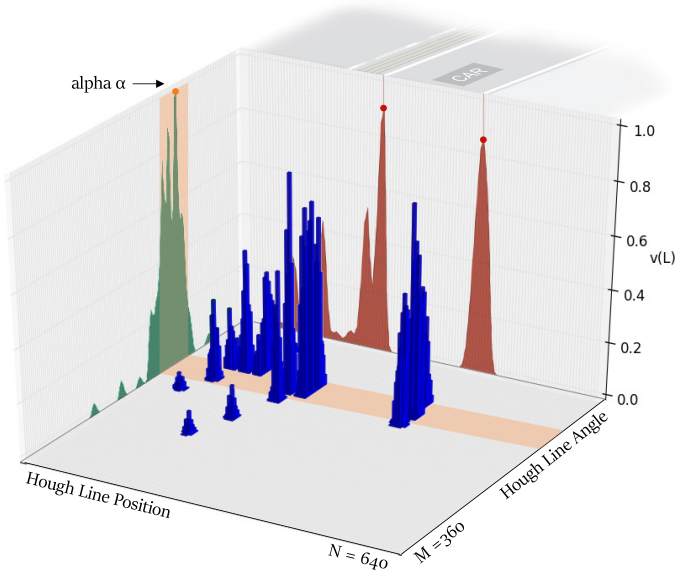

Figure 4: This figure illustrates the dominant angle calculation (Equation 3). $H_{x,y}^{2D}$ is represented by the blue bars. Green histogram represents $H_i^{1D}$ and $\alpha$ is its maximum. Used in further processing, the red histogram illustrates the projection of the selected Hough lines into the horizontal axis (Equation 8).

$$H_i^{1D} = \sum_{y=i-d}^{i+d} \sum_{x=0}^{N} H_{x,y}^{2D} \qquad (3)$$
$$\alpha = \max\{H_i^{1D} \mid 0 \leq i < M\}$$

In order to emphasize the locations where most strips are, the binary rotated region is projected (vertical maximum) into the horizontal axis, reducing it to a binary line. This binary line is then normalized to -1, 1. Crosswalks are assumed to have a distinguishable periodicity. Therefore, a convolution with its inverse should generate a minimum value in the center and maximum values as it shifts to the left or to the right. Based on this assumption, crosswalks are detected using a convolution of the projected crosswalk with its inverse. Considering $R$ as the result of this convolution, crosswalk detection for a projected region of size $n$ is defined by Equation 4 and illustrated in Figure 3 (b).

$$f(R) = \max\{R_0, \cdots, R_{\frac{n}{2}-1}\} + \max\{R_{\frac{n}{2}}, \cdots, R_n\})$$
$$\text{detected} = \begin{cases} crosswalk & \text{if } f(R) > 0 \\ none & \text{otherwise} \end{cases} \qquad (4)$$

### 2.3. Road Signs Detection and Classification

Road signs are visual marks added to the road to inform drivers about what is and what is not allowed in that pathway and to control its flow. Each country can implement its own symbols and use them based on their own rules, turning general detection and classification of road signs into a complex problem. This module combines a simple candidate extraction and classification procedure to detect and identify the different arrow types, the stop line, or any other "unknown" symbol (see figure Figure 5).

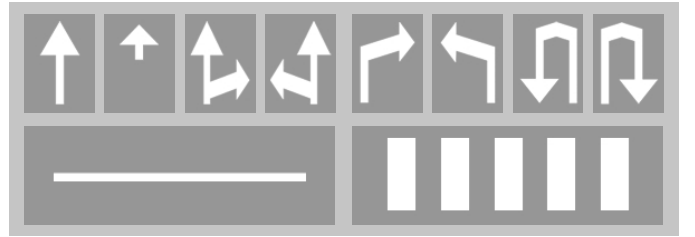

Figure 5: Road Signs and Crosswalk. Arrow signs in the upper row. Lower row: stop-line (left) and crosswalk (right). These symbols are defined in the Brazillian Manual of Traffic Signs [20].

*Candidates Extraction.* First of all, the input of this module is a combination of two feature maps (one for vertical and one for horizontal features) assumed to enhance pavement marking-like features (i.e., edges of markings). This combination is defined by $I^{DOG} \vee I^{VAD}$, where $\vee$ denotes a pixel-wise OR. Thereafter, if the previous lane position is known, lane region is corrected based on its angle. Subsequently, lane region is scanned. The scanning process performs a continuity check over the region's projection (horizontal maximum) into the vertical axis. As a result of this scanning process, a set of non-overlapping road sign candidates are extracted. These candidates are further filtered by a scanning process performed in their projection (vertical maximum) into the horizontal axis. Signs not having evidences in the center of the lane are ignored (road signs are assumed to be close to the center). The remaining signs are trimmed using the horizontal axis projection. In addition, pavement markings are expected to be brighter than asphalt. Therefore, the intensity of the sign candidates are compared with the intensity of the remaining surrounding region inside the lane. Finally, only candidates (signs) having average intensity higher than the average intensity plus one standard deviation of its surrounding region (asphalt) are considered.

*Classification.* Classification is done through a two-step process: stop line detection and template matching scheme. The detection of the stop line is performed through a proportion ($\frac{\text{width}}{\text{height}}$) threshold. Subsequently, a template matching scheme is applied in the remaining candidates. For this process, binary templates were generated using all 8 arrows from the Brazilian Manual of Traffic Sign [20] (see Figure 5). Candidates and patches are normalized to the same size ($32 \times 32$ pixels). Then, a template matching function, in this case Normalized Cross-Correlation [21], is calculated giving each candidate a set of coefficients $C = \{c_1, \cdots, c_n\}$, where $n = 8$ is the number of templates. Road sign candidates are classified into one of the arrows following Equation 5:

$$f(C) = \begin{cases} i & \text{if } \max\{C\} > \text{threshold} \\ unknown & \text{otherwise} \end{cases} \quad (5)$$

where $i$ is the index of the greatest coefficient, therefore $1 \leq i \leq 8$. More than one road sign can be detected in a given frame. The result of this module is a set of road signs classified into 1 of 10 classes: stop line, one of 8 arrows or unknown.

### 2.4. Crosswalk and Road Signs Removal

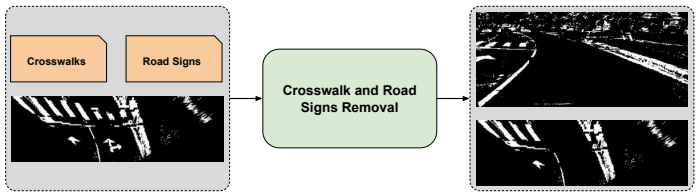

Figure 6: Previously detected crosswalks and road signs are given as input and this module removes them from the feature maps.

Although lane markings are meant to be visually distinguishable, roads have these other elements with similar properties that may confuse or degrade overall detection quality. Because at this point the presence of some of these symbols (i.e., crosswalk and road signs) is already known, they can be removed (Figure 6) from the feature maps used in further modules. This is expected to enhance lane markings detection, therefore increasing lane estimation accuracy.

### 2.5. Lane Estimation

The lane estimation process is a difficult problem for many reasons (e.g., lanes vary in shape, brightness, color and texture; traffic; etc.). In ELAS, a combination of Hough lines and Kalman filter is used for the base of the lane and a spline-based particle filter is used to model the curvature of the lane and distortions from the IPM. The estimation is performed in two phases because it is assumed that the lane base (i.e., lane region near the vehicle) can be linearly approximated, which simplifies the complexity of the particle filter used to model the lane as a spline. At first, individual lane candidates and lane measurements are generated based on Hough lines. Subsequently, Kalman is used to estimate the lane base. Finally, lane position (mainly curvature) is estimated by using a particle filter in which the lane estimation is a spline-based model.

#### 2.5.1. Lane Measurement Generation

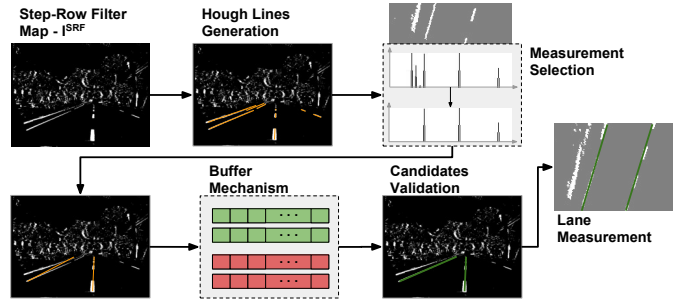

Figure 7: Overview of the lane measurement generation. The process receives $I^{SRF}$ as input and outputs a lane measurement (in green).

The goal of this task (Figure 7) is to generate a valid pair of Hough lines (i.e., lane measurement). It is assumed that most of the extracted Hough lines have the same orientation as the lane. Initially, a morphological skeleton [22] of $I^{SRF}$ is used to extract a set of lines (i.e., Hough lines) using Probabilistic Hough Transform [23]. This thinning is applied because large blobs tend to generate too many Hough lines with some of them misaligned with the lane markings. A Hough line can be represented as a set of points discretized in relation to the pixels in the vertical axis ($L = \{p_1, \cdots, p_n\}$). To remove outliers and properly choose a representative candidate for each lane, a 2D histogram ($H^{2D}$) is computed in the same way as in Subsection 2.2, but with a different bin counting. The contribution of each Hough line ($v(L)$) to the histogram is based on the number of evidences (i.e., white pixels in the feature map) under this line. Hough lines with evidences closer to it are preferred over those which evidences are farther, considering a maximum search length $b$. Therefore, $v(L)$ is calculated using Equation 6, where $dist(p)$ is the distance between $p$ and the closest evidence in the same horizontal line of $p$.

$$v(L) = \sum_{p \in L} \max\{0, b - dist(p)\} \quad (6)$$

Evidences are counted here, and in all the modules that uses the evidence counting process, on a combined map ($I^{CMB}$) defined by Equation 7.

$$I^{CMB} = I^{SRF} \wedge I^{INB} \quad (7)$$

where $\wedge$ is a pixel-wise AND operation.

Given this 2D histogram, the dominant angle $\alpha$ is calculated following Equation 3 using $d = 5$. Based on the dominant angle $\alpha$, $H^{2D}$ is masked to keep only Hough lines with angle closer than a threshold $\delta$, using $\delta = 15$. Then, masked 2D histogram is divided based on the vehicle's position into left and right histograms, and reduced to 1D normalized histograms using Equation 8.

$$H_x^{1D'} = \max\{H_{x,\alpha-\delta}^{2D}, \cdots, H_{x,\alpha+\delta}^{2D}\} \tag{8}$$

*Punishment Mechanism.* In the case of double lane markings, the innermost lane marking is expected to be chosen in the lane estimation process. Curbs and lane markings partially occluded by dust may also generate Hough lines in a certain amount that could lead to wrong estimates. To overcome this problem, a punishment mechanism was implemented. Basically, this mechanism punishes the outer Hough lines by using a punishment factor $\gamma$ on $H_x^{1D'}$. In fact, outer values are multiplied by $1 - (\gamma \times H_x^{1D'})$ in order to reduce them proportionally to inner values of $H_x^{1D'}$, resulting in $H^{1D}$. To avoid overweight of values around the center of the image, a neutral region in the center of the image (around the center of the vehicle) is created where this punishment is not applied. For each half (left and right) of $H^{1D}$, maximums are found. Each maximum has a Hough line associated with it. The output of this step is up to two Hough lines that generated those maximums in the current frame. Each Hough line can also be denoted as $L^c = \{\rho, \theta\}$, where $\rho$ is the position in the horizontal axis where a given candidate $L^c$ intersects the bottom of the image and $\theta$ is the direction angle of this line. Since the feature map contains noise, the output can be one of these: pair of independent lines; one of the lines, left or right; none of them.

*Buffer Mechanism.* Individual lane candidates are noisy. This mechanism acts as a temporal reinforcement factor. It ensures temporally coherent measures over previously accepted estimates. In real world, lane shapes are not expected to vary abruptly from one frame to another. Based on this fact, four buffers are created, two for each side: one for correct ($B^{correct}$) and another for incorrect ($B^{incorrect}$) candidates. The size of both buffers was empirically chosen: 10. A candidate is accepted into its $B^{correct}$ if buffer is not full or if it is close to the candidates in this buffer. A candidate $L^c = \{\rho, \theta\}$ is close to a buffer if the Euclidean distance between its $\rho$ and $\theta$ to the mean $\rho$ and $\theta$ of the elements $B^{correct}$ are less than a threshold (empirically defined as 15) respectively.

If a candidate for a side is accepted, its corresponding $B^{incorrect}$ is cleared. Otherwise, if a given candidate is rejected, it is stored into $B^{incorrect}$. If $B^{incorrect}$ is full, these sequentially rejected candidates are understood as a temporally coherent measure, and the current candidate should be accepted as correct. In this case, buffers are swapped in order to force a change on the estimates. Additionally, both, Kalman and particle filters (described later) are reset. This reset is performed to avoid smooth transitions in these swap situations that the estimation should adapt fast.

*Candidates Validation.* At this point, the following sets are possible: a pair of independent lines; one of the lines, left or right; none of them. Many scenarios might lead to missing line for one of the sides: true absence of one or both lane markings, a vehicle in front of the lane markings, noise, rejection by the buffer mechanism, etc. If at least one of the expected lane candidates is missing, its pair has to be generated by the system. A missing candidate is generated by projecting the only candidate available into the other side of the lane, using lane width from the previous frame. If both lane candidates are missing, the last estimation (if available) is used. Subsequently, they are combined (i.e., averaged) to compose a lane measurement. A measurement is represented by a base point, a lane width and an angle. This technique also reinforces the expected parallelism of lanes.

Finally, a lane measurement (i.e., a pair of lines) needs to be valid to be reported, otherwise it is reported as missing (i.e., none). The lane measurement is considered valid when $v(L^{left}) + v(L^{right})$ is greater than a threshold, where $v(L)$ is defined by Equation 6. There are 3 possible outputs for this step: lane measurement derived from a pair of lines, lane measurement derived from one line and the previous lane width, or no measurement.

### 2.5.2. Lane Base Estimation

Using the lane measurement given by the previous step, a Kalman filter [24] corrects the observations (i.e., lane measurements) based on a series of measurements over time. It acts like a smoother, since abrupt changes are unexpected most of the time, thus undesired. Nevertheless, in some cases, abrupt changes are desired (e.g., where there is no lane to be tracked, lane changes, etc.), but should not be confused with noise. Therefore, lane base estimative is controlled by a finite-state machine based on the presence or not of a lane measurement. The output of this module is a lane base (i.e., base point, lane width, and lane direction) estimated by the Kalman, or no estimative.

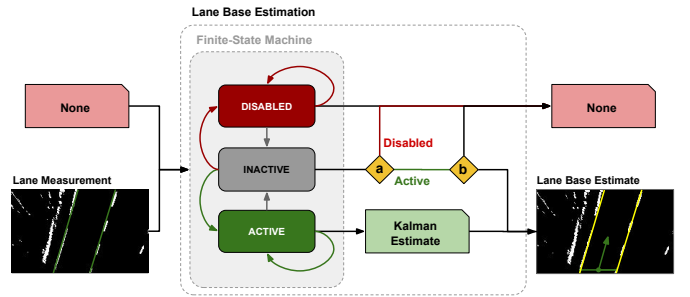

Figure 8: Lane base estimation overview. Lane measurements are put through a Finite-State Machine. The lane base estimative showed in dark green is given as output.

*Kalman Filter.* The Kalman is used to correct a lane measurement based on previous estimates. Lane measurement (a pair of lines, $L^{left}$ and $L^{right}$) is modeled in the observation vector as $z = \{p_b, p_t, w\}$, where $p_b$ and $p_t$ are intersections of the measurement with bottom and top of the image, respectively; and $w$ is the lane width.

*Finite-State Machine.* It works like a hysteresis, where Kalman is the controlled component of this finite-state machine (Figure 8). It controls the activation of the Kalman based on the presence or not of lane measurements. Basically, this state machine has 3 states:

- Active: if the measurement was derived from a pair of lines, it is passed to the Kalman to be predicted and corrected. If it was derived from one line (and the other side is just the projection based on the previous lane width), transition matrix of the Kalman is modified in order to ensure that the lane width is not updated (i.e., corrected). In both cases, the output is Kalman's corrected estimative and the filter remains active. If there was no measurement, the state is changed to inactive;

- Inactive: this is meant to be a transitory state, because when there were lane measurements for 10 sequential frames, Kalman should be activated again. Otherwise, if there were no lane measurements for the same amount of time, Kalman should be disabled. The output of this state is the same of the previous state (i.e., previous lane base estimate if it was active and no estimate if it was disabled, as shown in Figure 8 (a)). However, the previous lane base estimate must be valid to be reported (Figure 8 (b)). The lane base estimate is considered valid when $v(L^{left}) + v(L^{right})$ is greater than a threshold, where $L^{left}$ and $L^{right}$ are the lines derived (shown in yellow in Figure 8) from the current lane estimation, and $v(L)$ is defined by Equation 6;

- Disabled: the output of this state is no lane. Some situations are expected to lead to this state, such as: very unstable areas (e.g., lane markings constantly faded for a long period of time, intersections, etc.).

2.5.3. Lane Curvature Estimation

Previous modules rely on Hough lines to provide an initial estimative. Since it can not represent nonlinear models, this estimative is not suitable for curves. In this module, curvature is estimated using a spline-based particle filter. With the purpose of modeling curvature and adding robustness against deformations in the IPM, a lane estimative (i.e., particle) is defined. The quality of a particle is measured as a combination of evidences of lane marking and cleanness in the lane center. One particular issue with lane estimation is when there is a limited trustworthy zone ahead of the vehicle for estimating a lane (e.g., a car in front of the vehicle or a road intersection). In such cases, particle's quality should not be influenced by what is beyond this area. Therefore, the result of this module is a final lane estimation, bounded by a trustworthy area. This lane estimation is a virtual particle resulting from the weighted average of the current set of particles of the filter.

*Trustworthy Area.* ELAS considers detected stop lines as a delimiter of such area. Also, a search is performed in the middle of the previous lane estimation looking for evidences in $I^{VAD}$ (i.e., cars, motorcycles, or other obstacles are expected to produce evidences on this map). If there are evidences and they do not belong to one of the recognized road signs — except stop lines — they are taken as delimiters too. To apply this technique, the search space in the weight function is bounded to the vertical axis of this delimiter, since it defines a region to be considered by the particles of the filter.

*Particle Filter.* ELAS uses a particle filter [25] in order to estimate the curvature of the lane. Therefore, lane estimates are defined by two lane widths, to be robust against deformations in the IPM; and three control points uniformly distributed in the vertical axis, based on [16]. Then, a particle is defined as $\{w_1, w_2, x_1, x_2, x_3\}$. Nonetheless, the dimensionality of the particle filter is only 3, because $w_1$ and $x_1$ (lane width and position near vehicle, respectively) are calculated by the Kalman and not estimated by the particle filter. Only $w_2$, $x_2$ and $x_3$ are estimated by the filter, and $x_2$ and $x_3$ are sufficient as control points because their vertical positions are kept fixed (i.e., control points are uniformly distributed in the vertical axis).

Particle filter is a predictor-corrector method. Particles are randomly initialized around the lane measurement, using its position and direction. In the prediction phase, the particle filter aims to estimate a lane in the actual frame based on a previous set of particles. In our setup, a particle can move $x_2$ and $x_3$ randomly based on the direction of the lane, derived from the lane base estimate, and lane width $w_2$ can vary randomly with fixed standard deviation. In the correction phase, each particle has a weight assigned to it. This weight is correlated to its survival probability on the re-sampling process. At this point, all estimated particles have the same weight. Firstly, each particle has its weight ($W$) updated according to Equation 11, where each spline derived from the particle can be represented as a set of points discretized in relation to the pixels in the vertical axis ($S = \{p_1, \cdots, p_h\}$), bounded by the trustworthy area height $h$. Basically, this weight function gives higher weights to those particles with more evidences under their corresponding lane markings (Equation 9) and with a cleaner lane space between both sides of the estimated lane (Equation 10), where $\mu = 0$, $\sigma_1 = 1/3$, $\sigma_2 = 1/12$ and $G(x \mid \mu, \sigma_1)$ is the Gaussian function.

$$l = \frac{1}{h} \sum_{p \in S^{left}} I_p^{CMB} \text{ and } r = \frac{1}{h} \sum_{p \in S^{right}} I_p^{CMB}$$

$$W_1' = 1 - \left( l \times r + (1 - l \times r) \left( \frac{l+r}{2} \right) \right) \quad (9)$$

$$W_2' = \sum_{p \in S^{left}} \sum_{j=0}^{d} I_{p(x+j,y)}^{CMB} + \sum_{p \in S^{right}} \sum_{j=0}^{d} I_{p(x-j,y)}^{CMB} \quad (10)$$

$$W = G(W_1'|\ \mu, \sigma_1) \times G(W_2'|\ \mu, \sigma_2) \quad (11)$$

After updating the weight of each particle, a resampling technique is used. ELAS uses the Low Variance Sampling [26] to resample the particles based on their weights: higher weights mean higher chances to be resampled. Final lane estimation is a spline-based model represented by a virtual particle that is generated based on the weighted average of the current set of particles. For more details on the particle filter, see Figure 9.

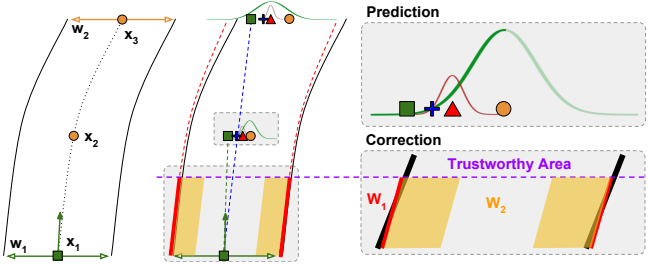

Figure 9: Particle filter overview. A particle ($\{w_1, w_2, x_1, x_2, x_3\}$) is illustrated on the left, where the components $w_1$ and $x_1$ (in green) are defined by the Kalman in the previous module and the components $w_2$, $x_2$ and $x_3$ (in orange) are estimated by the filter. The Prediction and Correction steps are illustrated on the right. In the prediction, $x_2$ (blue cross) is randomly chosen following a combination of gaussian distributions as follows. A projection (green square) of the base point $x_1$ is calculated using the angle of the lane. A reference point (red triangle) is randomly chosen using a gaussian (in green) with a dynamic standard deviation and centered in the previous particle position (orange circle). The gaussian is mirrored to ensure an estimation in the direction of the green square. The standard deviation is calculated using 1/3 of the distance between the orange circle and the green square. The new location (blue cross) of the particle for the control point $x_2$ is finally randomly chosen from a fixed size normal distribution (in red) centered around the red triangle. The same procedure is applied to predict $x_3$. However, to allow for more mobility, the dynamic standard deviation is not divided by 3. Additionally, the base point $x_1$ is not projected with the Kalman estimated angle, but in the direction of the newly estimated $x_2$ (blue cross of the middle). In the Correction step, evidences under the estimated lanes (red lines) and the evidences in the inner space of the lane (orange regions) are counted (Equation 9 and Equation 10, respectively) considering the trustworthy area.

### 2.5.4. Lane Departure Measure

To enable lane departure warning (LDW), ELAS estimates the position of the vehicle related to the host lane. It assumes a forward-looking camera mounted on the top of this vehicle. The car position is expected to be in the center of the image. Using the known car position and lane center estimation, ELAS can calculate vehicle's position related to the center of the lane.

### 2.6. Lane Marking Type (LMT) Classification

Each lane side is classified into one of 7 types of LMTs (Figure 10). The lane estimation is derived into both sides, left and right lanes, and a region along the lane is analyzed considering the evidences in the $I^{INB}$ map. Firstly, the number of yellow lane marking evidences is counted. The count is based on whether the pixel value is within a specific range (($30°, 31\%, 31\%$) and ($50°, 78\%, 78\%$)) in the HSV color-space. If less than half of the evidences under the lane are considered yellow, than the lane is assumed to be one of the 2 white LMTs (Figure 10 (a)). Otherwise, it is assumed to be one of the 5 yellow LMTs. Subsequently, two counts are made: one for calculating the percentage of evidences to distinguish between "dashed" or "solid"; and, one to distinguish between "single" or "double". The counts are made in the $I^{INB}$ map considering a region along the lane. The first counts the number of evidences along lane considering the horizontal maximum projection for each point of the lane (i.e., a point in the lane has an evidence if any of its horizontal neighbors has). The second counts the number of horizontal "white-black-white" (WBW) patterns for each point of the lane (i.e., a point in the lane has a WBW if there was a white to black to white transition along the horizontal axis). White lanes with less than 30% evidences are assumed WSD, otherwise it is assumed WSS (Figure 10 (b)). Yellow lanes with less than 20% WBW patterns are assumed "single" (Figure 10 (c)) and are further classified in YSS and YSD using the same procedure of the white lanes (Figure 10 (b)). Otherwise, they are assumed "double" and need to be further classified between "mixed" and "solid" (Figure 10 (d)). Lane is assumed YDS if it has more than 80% of WBW patterns, otherwise it is assumed "mixed", and need to be further classified between YMS and YMD (Figure 10 (e)). This distinction is made by analyzing the portions of the lane without WBW patterns to identify which side has the solid line.

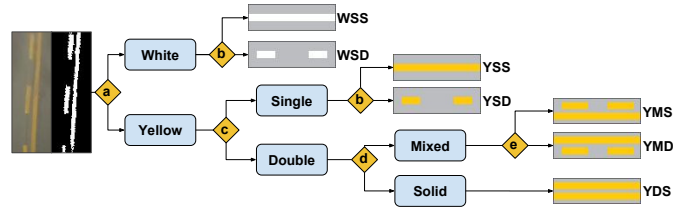

Figure 10: Lane marking types (LMT) as detected and classified according to a threshold-based method. These are the LMTs: white-single-solid (WSS), white-single-dashed (WSD), yellow-single-solid (YSS), yellow-single-dashed (YSD), yellow-mixed-solid (YMS), yellow-mixed-dashed (YMD) and yellow-double-solid (YDS). Decision processes (a, b, c, d, and e) are described in Subsection 2.6.

To avoid unstable reports caused by noisy lane markings (fading, covered by dust, occluded by an obstacle or even during transition areas between different LMTs), a buffer mechanism was added. In these buffers (size 30), a winner-takes-all approach is used to report the LMT. The output of this module is a pair of LMT, one for the left and another for the right lane side.

### 2.7. Adjacent Lane Detection

Based on the final lane estimation and classified LMTs, ELAS looks for immediate adjacent lanes, i.e., one lane to the left and one to the right of the ego-lane. According to the Brazilian Manual of Traffic Sign, only "white single solid" (WSS) lane markings may not have an adjacent lane. Therefore, this is the only case the system needs to take a decision. In order to decide about the presence or not of adjacent lanes in these cases, a searching area taking the estimated lane width in account is considered. Given this area, the system checks two conditions (Figure 11): if there are Hough lines with angle close to the ego-lane angle (i.e., less than $\delta$ of difference, where $\delta = 15$) in the expected area of the adjacent lane; and if the space between this area and the ego-lane has a very small amount of Hough lines. For each lane side individually, when LMT is WSS and if both conditions are met, this module reports the existence of such adjacent lane based on this process. Otherwise, if LMT is other than WSS, the existence of an adjacent lane is presumed, based on the Brazilian Manual of Traffic Sign [20]. The lane marking color allows for deciding between same or opposite direction lane.

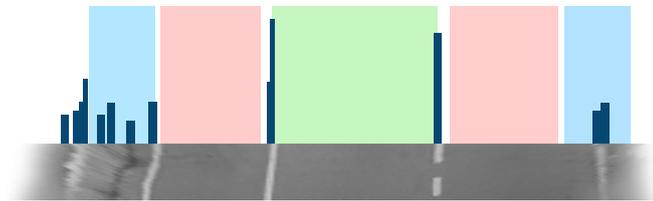

Figure 11: Adjacent Lane Detection. The dark blue bars shows the histogram counts for the Hough lines generated by the figure at the bottom of the image. The green region is the estimated lane region. The blue region is the searching region calculated based on the lane width of the current estimation. The red region is the space between between the ego-lane and the adjacent lane.

## 3. Materials and Experimental Methodology

In order to validate our system, we ran experiments using a novel dataset. Besides common performance measurements, a set of experiments were conducted to analyze the impact of some unexpected setups (e.g., low processing power and low image quality).

### 3.1. Database

To validate our system where it is expected to run, we recorded and annotated a novel dataset. This dataset was recorded using a GoPRO HERO 3 camera in Brazil in different days at 29.97 frames per second. It comprises more than 20 scenes (i.e., more than 15,000 frames) containing all sorts of challenges we expect to encounter in the real world: highways, urban road, traffic, intersections, lane changes, writings on the road, shadows, wobbly capture, different weather, slightly different camera positions, faded lane markings, multiple LMT transitions, different vehicle speed, etc. Scenes were recorded in three cities: Vitória, Vila Velha, and Guarapari on the state of Espírito Santo, Brazil. Each frame has 640 × 480 pixels. This dataset was manually annotated for relevant events to the research community, such as: lane estimation, change, and centering; road markings; intersections; LMTs; crosswalks; and adjacent lanes. Lane position ground truth generation was done following [27]. During LMT transitions, the same strategy of [28] was adopted, where both types were annotated. The dataset will be made publicly available[1].

### 3.2. Metrics

The metrics we used to measure the performance of the lane estimation are closely related to some of the metrics proposed in [29]. To evaluate the lane estimation, four evaluation points were chosen. These points are equally distributed in the region of interest in respect to their distances from the car. In this way, the three initial points are considered as near and the last one as far. Errors near the car are assumed to have more impact in the effectiveness of the systems that rely on this information. In this perspective, the mean absolute error of the lane estimation is reported. This error is the difference in pixels of the estimated position and the ground truth. In contrast to the *Lane Position Deviation* proposed in [29], we reported the error in terms of percentage of the lane width. In addition, their metric is a combination of the errors in both, near and far depth of view, while ours are reported separately. Another metric is the Lane Center Deviation. It comprises the difference of the lane center predicted by our system and the lane center of the ground truth. This metric is very similar to the *Ego-vehicle localization* proposed in [29], and the differences are the same as the previous metric. The mean absolute error of the lane center deviation is also reported in terms of the percentage of the lane width. Additionally, execution time is reported in milliseconds. The results of the experiment performed for the analysis of the execution time (varying the complexity of the system) resembles the *Computation Efficiency and Accuracy* metric proposed in [29]. Finally, for the detection tasks (i.e., crosswalk, pavement signs, lane marking types and adjacent lanes), accuracy is reported.

### 3.3. Experiments

Each module was evaluated to measure their performances and understand their behaviors in various scenarios.

---

[1]www.lcad.inf.ufes.br/wiki/index.php?title=Ego-Lane_Analysis_System

*Kalman with/without Particle Filter.* Roughly, lane estimation is performed in two steps in our system: Hough-based and particle filter-based steps. The former is required, but it is intrinsically limited in curve roads. The latter comes to fulfill this gap, estimating the curvature and giving robustness against the fixed ground plane assumptions that naturally fails sometimes. On the other hand, the latter also requires more processing capacity. This experiment consists on measuring the impact of the particle filter, analyzing the trade-off involved with the use of the particle filter and checking if it is viable to rely only in the Hough-based output. As particle filter overall performance is directly related to the number of particles, we also experimented with different number of particles {50, 100, 200, 400}.

*Image quality.* Our dataset was annotated in $640 \times 480$ pixels and was captured using a high quality camera. However, this is not always the case, and this experiment measures ELAS performance with lower quality images. To simulate lower quality capture, we down-sampled frames to $480 \times 360$ (75%), $320 \times 240$ (50%) and $160 \times 120$ (25%) using nearest-neighbor as the interpolation method and up-sampled back to $640 \times 480$ using a linear interpolation. This process causes loss of data (Figure 12). These experiments aims to test robustness against lower quality images without having to capture or annotate a dataset all over again.

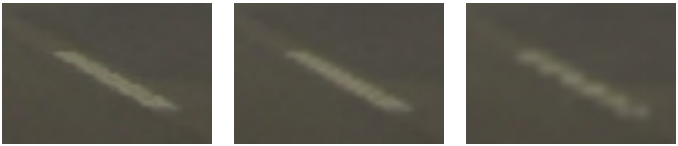

Figure 12: With this technique, quality loss causes a degradation of the image, in special lane markings. From left to right: 75%, 50% and 25%.

*Frame rate.* Our dataset was captured at 29.97 frames per second (FPS). Nevertheless, the system might need to operate with lower frame rates due to a limitation in the camera or to limited processing power. To simulate lower frame rates, some of the frames were dropped in order to achieve 20, 15, and 10 frames per second.

*3.4. Setup*

All experiments were performed on a desktop with Intel Core i7-4770 (3.40GHz) and 16GB RAM. ELAS was implemented in C++ using the open source library OpenCV. Despite this setup, ELAS used up to 15% of the processing power and up to 40MB of RAM during experiments.

## 4. Results and Discussion

The performance and accuracy impact of both scenarios (with and without particle filter) can be seen in Figure 13. As expected, when the particle filter is disabled (i.e., only Kalman output is being used and reported), there is a decrease in the execution time. As it can be seen, Kalman-only presents lower accuracy in the far region, where curvature happens. Additionally, a higher number of particles increases particle filter accuracy at the cost of an increase in execution time. As all these configurations allow real-time execution (more than 30 frames per second), the best was chosen (PF400) to benchmark other experiments. PF400 configuration achieved mean absolute error of 1.3% ($\pm$ 0.2) in the near region and 3.6% ($\pm$ 0.3) in the far region executing in 27.2ms ($\pm$ 2.9).

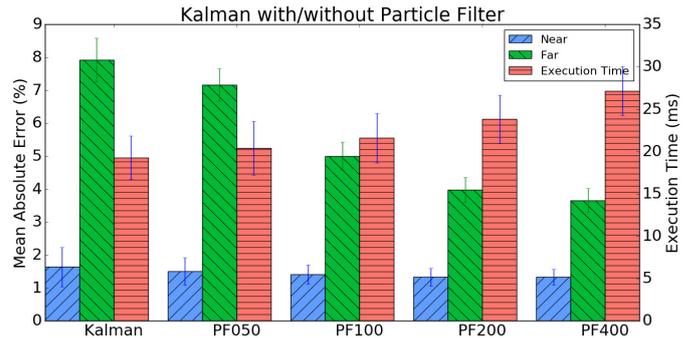

Figure 13: The mean absolute error (%) of the near and far regions are reported. Additionally, execution time (ms) is shown. In this case, particle filter was tested with 50, 100, 200 e 400 particles (PF50, PF100, PF200 and PF400, respectively).

ELAS achieves an average of 86.7% ($\pm$ 2.8) in the classification accuracy of road signs and crosswalks. As an example of misclassification, a crosswalk was being intermittently classified between "unknown" and crosswalk because of a crosswalk variation containing arrows. Our LMT detector overall classification rate was 93.1% ($\pm$ 1.9), with the lowest accuracy in the rainy scene. It was also noted that, usually, LMT misclassifications are from yellow-double to yellow-single-solid and white-dashed to white-solid. This is less dangerous than misclassifying in the opposite direction, since these cases can lead control systems to take dangerous decisions (e.g. allow a lane change while it should not). Also, one of the advantages of our approach is that it does not need training [28]. On the other hand, one limitation is the color detection in cases where overall color is expected to be changed (e.g. snow and dusty windshield in a sunny day). By design, our adjacent lane detector performance is constrained by the accuracy of our LMT detector. Therefore, the overall adjacent lane detection rate was 89.4% ($\pm$ 2.6). The common problems of this detector are correlated to occlusions of the adjacent lane markings. Qualitative results can be seen in Figure 17. A demonstration video of ELAS is also available[2]. In the video, ELAS is also executed in a sequence with more than 20km, for qualitative analysis.

We also wanted to validate our system in other setups. Due to the prohibitive cost of generating the database again, we used some techniques to simulate these setups. The first one was lower frame rate (i.e., frames per second, FPS). As expected, lower FPS causes lower accuracy (Figure 14), but ELAS maintains reasonable estimation quality even on 10 FPS.

---

[2] https://www.youtube.com/watch?v=NPU9tiyA8vw

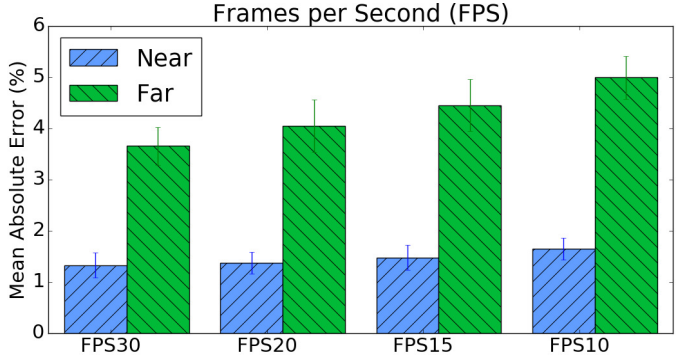

Figure 14: Frames per Second. ELAS was tested with 30, 20, 15, and 10 frames per second (FPS30, FPS20, FPS15, FPS10, respectively). All these tests were performed using PF400. In fact, FPS30 was reported again for completeness, because it is the same as PF400.

The second one was lower image quality (Figure 15). Based on the technique described earlier, some noise was added to the image as if the image had came from a lower resolution camera or a lower quality one. As expected, lower quality do cause a decrease in the overall accuracy, but again the system was capable of maintaining a good estimation quality.

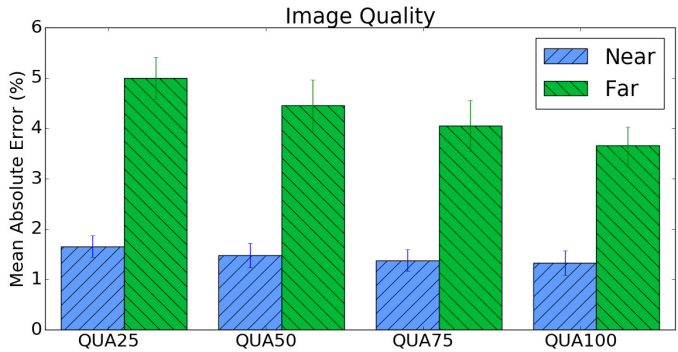

Figure 15: Image Quality. Lower image qualities were simulated based on the image resolutions where 100% (QUA100) represents the original image and the others represent the relative resolutions: 75%, 50%, and 25% (QUA75, QUA50, QUA25, respectively).

Lane center deviation is accurately calculated, with an absolute mean error of 0.9% (± 0.18). Based on this lane center deviation, all lane changes are detected. Execution time varies according to the amount of elements to process in a given frame. As reported in Table 1, ELAS devotes most of its pipeline to lane estimation (Hough lines + Kalman + particle filters). Even though, the system runs in real-time (33+FPS) considering the used experiment setup. Overall execution time can be further improved by parallel implementation, benefiting from the modular approach.

| Execution Time (ms) | |
|---|---|
| Feature Maps Generation | 2.64 ± 0.27 |
| Crosswalk and Road Signs Detection and Removal | 1.59 ± 0.87 |
| Candidates Generation and Kalman | 8.46 ± 0.76 |
| Particle Filter | 7.57 ± 0.55 |
| Lane Marking Type Detection | 6.91 ± 0.44 |
| Adjacent Lane Detection | 0.03 ± 0.01 |
| **Total** | **27.2 ± 2.90** |

Table 1: Execution Time Performance of ELAS with 400 particles (in ms)

Finally, ELAS was compared with [16] in the newly proposed dataset. As it can be seen in Figure 16, ELAS lane estimation outperforms [16].

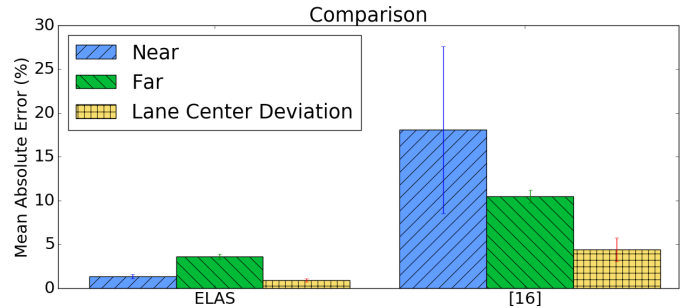

Figure 16: Comparison between ELAS and [16]

## 5. Conclusion

This paper presents a system (ELAS) for ego-lane analysis. Additionally, a novel dataset with more than 20 scenes (15,000+ frames) was generated, manually annotated and made publicly available to enable future fair comparisons. Also, the implementation of ELAS is publicly available.

Our experiments evaluated ELAS using real-world driving environments. Results showed our system presents high performance in lane estimation and high classification rates on the other examined tasks (i.e., road signs, crosswalks, LMT and adjacent lanes). Additionally, lane center deviation results proved the system can be used to perform lane departure warning accurately. Moreover, experimental results validated its robustness against lower quality and lower frame rates. Ultimately results showed that ELAS is able to robustly perform real-time ego-lane analysis.

## Acknowledgment

The authors would like to thank Fundação de Amparo Pesquisa do Espírito Santo FAPES (grants 53631242/11

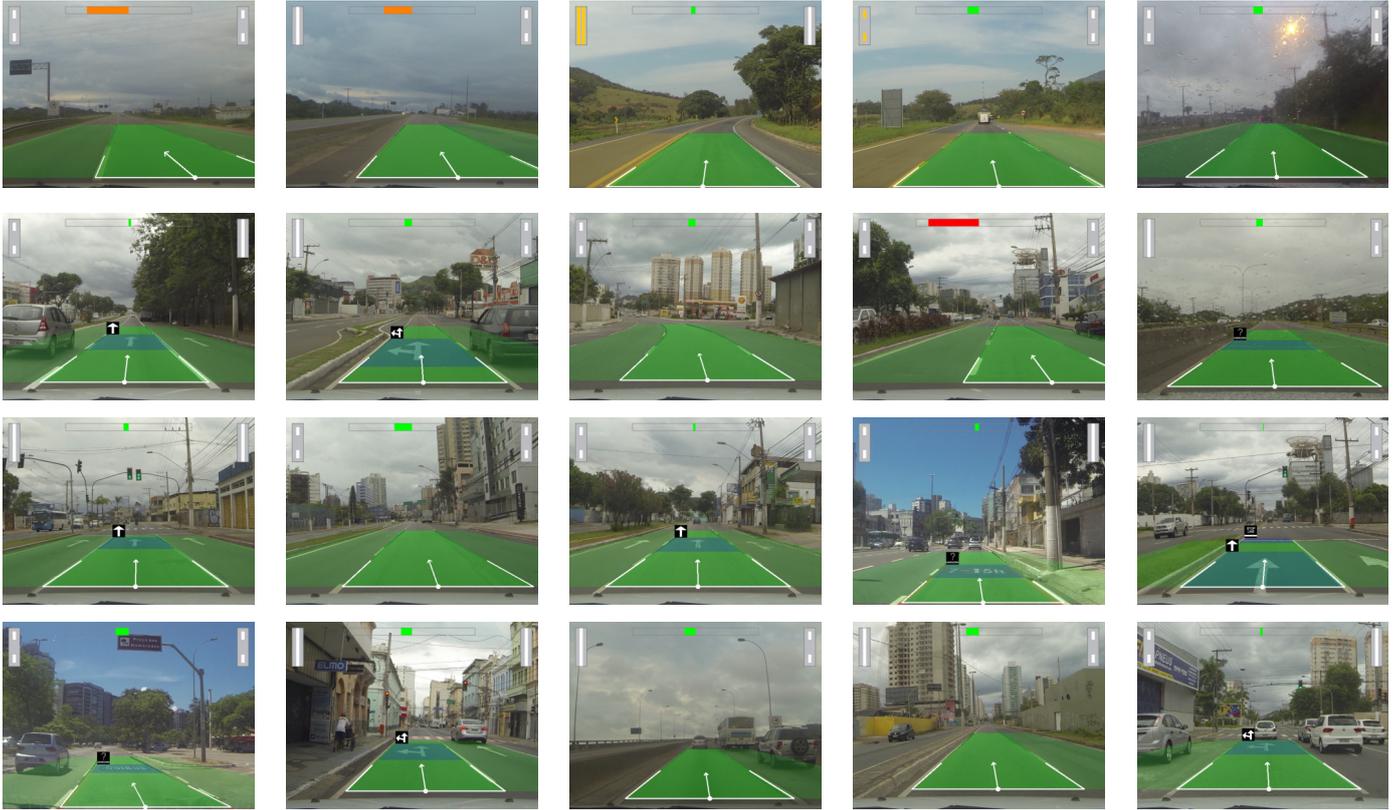

Figure 17: Qualitative results of ELAS for some of the scenes. First four columns show correct outputs while last column shows some of the incorrect results.

and 60902841/13, and scholarship 66610354/2014), Coordenação de Aperfeiçoamento de Pessoal de Nível Superior CAPES (grant 11012/13-7), and Conselho Nacional de Desenvolvimento Científico e Tecnológico CNPq, Brazil (grants 552630/2011-0, 312786/2014-1) for their financial support to this research work.